# Greedy Algorithms for Sparse Reinforcement Learning

**Christopher Painter-Wakefield** PAINT007@CS.DUKE.EDU
**Ronald Parr** PARR@CS.DUKE.EDU
Duke University, Durham NC

## Abstract

Feature selection and regularization are becoming increasingly prominent tools in the efforts of the reinforcement learning (RL) community to expand the reach and applicability of RL. One approach to the problem of feature selection is to impose a sparsity-inducing form of regularization on the learning method. Recent work on $L_1$ regularization has adapted techniques from the supervised learning literature for use with RL. Another approach that has received renewed attention in the supervised learning community is that of using a simple algorithm that greedily adds new features. Such algorithms have many of the good properties of the $L_1$ regularization methods, while also being extremely efficient and, in some cases, allowing theoretical guarantees on recovery of the true form of a sparse target function from sampled data. This paper considers variants of orthogonal matching pursuit (OMP) applied to reinforcement learning. The resulting algorithms are analyzed and compared experimentally with existing $L_1$ regularized approaches. We demonstrate that perhaps the most natural scenario in which one might hope to achieve sparse recovery fails; however, one variant, OMP-BRM, provides promising theoretical guarantees under certain assumptions on the feature dictionary. Another variant, OMP-TD, empirically outperforms prior methods both in approximation accuracy and efficiency on several benchmark problems.

## 1. Introduction

Feature selection and regularization are becoming increasingly prominent tools in the efforts of the reinforcement learning (RL) community to expand the reach and applicability of RL (Parr et al., 2007; Mahadevan & Maggioni, 2007; Johns, 2010; Johns et al., 2010; Ghavamzadeh et al., 2011). Very often (though not always (Farahmand et al., 2008)) sparseness is viewed as a desirable goal or side effect of regularization. If the true value function is known to be sparse, then the reasons for desiring a sparse solution are clear. Even when the form of the true value function is not known, sparsity may still be desired because sparsity can act as a regularizer, and because sparse solutions tend to be more understandable to humans and more efficient to use. Favoring sparsity can lead to faster algorithms in some cases (Petrik et al., 2010).

One optimization-based approach to the problem of feature selection is to impose a sparsity-inducing form of regularization on the learning method. Recent work on $L_1$ regularization has adapted techniques from the supervised learning literature (Tibshirani, 1996) for use with RL (Kolter & Ng, 2009). Another approach that has received renewed attention in the supervised learning community is that of using a simple algorithm that greedily adds new features (Tropp, 2004; Zhang, 2009). Such algorithms have many of the good properties of the $L_1$ regularization methods, while also being extremely efficient and, in some cases, allowing theoretical guarantees on recovery of the true form of a sparse target function from sampled data despite the myopia associated with greediness.

The most basic greedy algorithm for feature selection for regression, matching pursuit, uses the correlation between the residual and the candidate features to decide which feature to add next (Mallat & Zhang, 1993). This paper considers a variation called orthogonal matching pursuit (OMP), which recomputes the residual after each new feature is added, as applied to reinforcement learning. It is related to BEBFs (Parr et al., 2007), but it differs in that it selects features from a finite dictionary. OMP for RL was explored by Johns (2010) in the context of PVFs (Mahadevan & Maggioni, 2007) and diffusion wavelets (Mahadevan & Maggioni, 2006), but aside from this initial exploration of the topic, we are not aware of any efforts to bring the theoretical and empirical understanding of OMP for reinforcement learning to parity with the understanding of OMP as applied to regression.

This paper contributes to the theoretical and practical understanding of OMP for RL. Variants of OMP are analyzed and compared experimentally with existing $L_1$ regularized approaches. We demonstrate that perhaps the most natural scenario in which one might hope to achieve sparse recovery fails; however, one variant, OMP-BRM, provides promising theoretical guarantees under certain assumptions on the feature dictionary. Another variant, OMP-TD, lacks theoretical guarantees but empirically outperforms OMP-BRM and prior methods both in approximation accuracy and efficiency on several benchmark problems.

## 2. Framework and Notation

We aim to discover exact or good, approximate value functions for Markov reward processes (MRPs): $M = (S, P, R, \gamma)$. Given a state $s \in S$, the probability of a transition to a state $s' \in S$ is given by $P(s'|s)$, and results in an expected reward of $R(s)$. We do not address the question of optimizing the policy for a Markov Decision Process, though we note that policy evaluation, where $P = P_\pi$, by some policy $\pi$, is an important intermediate step in many algorithms. A discount factor $\gamma$ discounts future rewards such that the present value of a trajectory $s_{t=0} \dots s_{t=n}$ is $\sum_{t=0}^{n} \gamma^t R(s_t)$.

The true value function $V^*$ over states satisfies the *Bellman equation*:

$$V^* = TV^* = R + \gamma P V^*,$$

where $T$ is the *Bellman operator* and $V^*$ is the fixed point of this operator.

In practice, the value function, the transition model, and the reward function are often too large to permit an explicit, exact representation. In such cases, an approximation architecture is used for the value function. A common choice is $\hat{V} = \Phi w$, where $w$ is a vector of $k$ scalar weights and $\Phi$ stores a set of $k$ features in an $n \times k$ feature matrix. Since $n$ is often intractably large, $\Phi$ can be thought of as populated by $k$ linearly independent *basis functions*, $\varphi_1 \dots \varphi_k$, which define the columns of $\Phi$. We will refer to a basis formed by selecting a subset of the features using an index set as a subscript. Thus, $\Phi_\mathcal{I}$ contains a subset of features from $\Phi$, where $\mathcal{I}$ is a set of indices such that basis function $\varphi_i$ is included in $\Phi_\mathcal{I}$ if $i \in \mathcal{I}$.

For the purposes of estimating $w$, it is common to replace $\Phi$ with $\hat{\Phi}$, which samples rows of $\Phi$, though for conciseness of presentation we will use $\Phi$ for both, since algorithms for estimating $w$ are essentially identical if $\hat{\Phi}$ is substituted for $\Phi$. A number of linear function approximation algorithms such as LSTD (Bradtke & Barto, 1996) solve for the

**Algorithm 1** OMP

**Input:** $X \in \mathbb{R}^{n \times k}, y \in \mathbb{R}^n, \beta \in \mathbb{R}$.
**Output:** Approximation weights $w$.
$\mathcal{I} \leftarrow \{\}$
$w \leftarrow 0$
**repeat**
  $c \leftarrow |X^T(y - Xw)|$
  $j \leftarrow \arg\max_{i \notin \mathcal{I}} c_i$
  **if** $c_j > \beta$ **then**
    $\mathcal{I} \leftarrow \mathcal{I} \cup \{j\}$
  **end if**
  $w_\mathcal{I} \leftarrow X_\mathcal{I}^\dagger y$
**until** $c_j \le \beta$ **or** $\bar{\mathcal{I}} = \{\}$

$w$ which is a fixed point:

$$\Phi w = \Pi^\sigma (R + \gamma \Phi' w) = \Pi^\sigma T \Phi w, \tag{1}$$

where $\Pi^\sigma$ is the $\sigma$-weighted $L_2$ projection and where $\Phi'$ is $P\Phi$ in the explicit case and is composed of sampled next features in the sampled case. We likewise overload $T$ for the sampled case.

We make use of the notation $\Phi^\dagger$ to represent the pseudo-inverse of $\Phi$ specifically defined as $\Phi^\dagger \equiv (\Phi^T\Phi)^{-1}\Phi^T$. In general we assume that the weighting function $\sigma$ is implicit in the sampling of our data, in which case the projection operator above is simply $\Pi = \Phi\Phi^\dagger$.

We also consider algorithms which solve for $w$ minimizing the Bellman error:

$$w = \arg\min_w \|T\Phi w - \Phi w\|. \tag{2}$$

This is the Bellman residual minimization (BRM) approach espoused by Baird (1995).

## 3. Prior Art

### 3.1. OMP for Regression

Algorithm 1 is the classic OMP algorithm for regression. It is greedy in that it myopically chooses the feature with the highest correlation with the residual and never discards features. We say that target $y$ is *m-sparse in* $X$ if there exists an $X_{opt}$ composed of $m$ columns of $X$ and corresponding $w_{opt}$ such that $y = X_{opt} w_{opt}$, and $X_{opt}$ is minimal in the sense there is no $X'$ composed of fewer columns of $X$ which can satisfy $y = X'w$. Of the many results for sparse recovery, Tropp's (2004) is perhaps the most straightforward:

**Theorem 1** *If $y$ is m-sparse in $X$, and*

$$\max_{i \notin opt} \|X_{opt}^\dagger x_i\|_1 < 1,$$

*then OMP called with $X$, $y$, and $\beta = 0$ will return $w_{opt}$ in $m$ iterations.*

In the maximization, the vectors $x_i$ correspond to the columns of $X$ which are not needed to reconstruct $y$ exactly. In words, this condition requires that the projection of any suboptimal feature into the span of the optimal features must have small weights. Note that any orthogonal basis trivially satisfies this condition.

Tropp also extended this result to the cases where $y$ is approximately sparse in $X$, and Zhang (2009) extended the result to the noisy case.

### 3.2. $L_1$ Regularization in RL

In counterpoint to the greedy methods based on OMP, which we will explore in the next section, much of the recent work on feature selection in RL has been based on least-squares methods with $L_1$ regularization. For regression, Tibshirani (1996) introduced the LASSO, which takes matrix $X$ and target vector $y$ and seeks a vector $w$ which minimizes $\|y - Xw\|^2$ subject to a constraint on $\|w\|_1$. While Tibshirani uses a hard constraint, this is equivalent to minimizing

$$\|y - Xw\|^2 + \beta\|w\|_1, \tag{3}$$

for some value of $\beta \in \mathbb{R}^+$.

Loth et al. (2007) apply the LASSO to Bellman residual minimization. Replacing the residual $y - Xw$ in equation 3 with the Bellman residual, we obtain

$$\|R + \gamma\Phi' w - \Phi w\|^2 + \beta\|w\|_1.$$

Trivially, if we let $X = \Phi - \gamma\Phi'$ and $y = R$, we can substitute directly into equation 3 and solve as a regression problem. The regression algorithm used by Loth et al. is very similar to LARS (Efron et al., 2004).

A harder problem is applying $L_1$ regularization in a fixed point method akin to LSTD. The $L_1$ *regularized linear fixed point* is the vector $w$ solving

$$w = \arg\min_u \|R + \gamma\Phi' w - \Phi u\|^2 + \beta\|u\|_1.$$

introduced by Kolter & Ng (2009). Kolter and Ng provide an algorithm, LARS-TD, which closely follows the approach of LARS. Johns et al. (2010) followed LARS-TD with an algorithm, LC-TD, which solves for the $L_1$ regularized linear fixed point as a linear complementarity problem.

It is instructive to note that LARS bears some resemblance to OMP in that it selects as new features those with the highest correlation to the residual of its current approximation. However, where OMP is purely greedy and seeks to use all active features to the fullest, LARS is more moderate and attempts to use all active features *equally*, in the sense that all active features maintain an equal correlation with the residual. One aspect of the LARS approach that sets it quite apart from OMP is that LARS will remove features from the active set when necessary to maintain its invariants.

Intuitively, LARS and algorithms based on LARS such as LARS-TD have an advantage in minimizing the number of active features due to their ability to remove features. LC-TD also adds and removes features. These methods do suffer from some disadvantages related to this ability, however. LARS-TD can be slowed down by the repeated adding and removing of features. Worse, both LARS-TD and LC-TD involve computations which are numerically sensitive and are not guaranteed to find the desired solution in all cases since (unlike in the pure regression case) the task of finding an $L_1$ regularized linear fixed point is not a convex optimization problem.

**Algorithm 2** OMP-BRM

**Input:**
$\Phi \in \mathbb{R}^{n\times k} : \Phi_{ij} = \varphi_j(s_i)$,
$\Phi' \in \mathbb{R}^{n\times k} : \Phi'_{ij} = \varphi_j(s'_i)$,
$R \in \mathbb{R}^n : R_i = r_i$,
$\gamma \in [0, 1)$,
$\beta \in \mathbb{R}$
**Output:**
Approximation weights $w$.

**call** OMP with $X = \Phi - \gamma\Phi'$, $y = R$, $\beta = \beta$

## 4. OMP for RL

We present two algorithms for policy evaluation: OMP-BRM and OMP-TD.[1] As the names suggest, the first algorithm is based on Bellman residual minimization (BRM), while the second is based on the linear TD fixed point. Algorithm 2, OMP-BRM, is the simpler algorithm in the sense that it essentially performs OMP with features $\Phi - \gamma\Phi'$ using the reward vector as the target value. OMP-BRM is different from the OMP-BR algorithm introduced by Johns (2010), which selected basis functions from $\Phi$.

Algorithm 3, OMP-TD, applies the basic OMP approach to build a feature set for LSTD.[2] OMP-TD is similar in approach to the approximate BEBF algorithm of Parr et al. (2007), in which each new feature is an approximation to the current Bellman residual. In OMP-TD, rather than approximate the Bellman residual, we simply add the feature

[1] Note that both algorithms reduce to OMP when $\gamma = 0$.
[2] Johns (2010) called the same algorithm OMP-FP.

**Algorithm 3** OMP-TD

**Input:**
$\Phi \in \mathbb{R}^{n\times k} : \Phi_{ij} = \varphi_j(s_i)$,
$\Phi' \in \mathbb{R}^{n\times k} : \Phi'_{ij} = \varphi_j(s'_i)$,
$R \in \mathbb{R}^n : R_i = r_i$,
$\gamma \in [0,1)$,
$\beta \in \mathbb{R}$
**Output:**
Approximation weights $w$.

$\mathcal{I} \leftarrow \{\}$
$w \leftarrow 0$
**repeat**
  $c \leftarrow |\Phi^T(R + \gamma\Phi' w - \Phi w)|/n$
  $j \leftarrow \arg\max_{i\notin\mathcal{I}} c_i$
  **if** $c_j > \beta$ **then**
    $\mathcal{I} \leftarrow \mathcal{I} \cup \{j\}$
  **end if**
  $w_{\mathcal{I}} \leftarrow (\Phi_{\mathcal{I}}^T\Phi_{\mathcal{I}} - \gamma\Phi_{\mathcal{I}}^T\Phi'_{\mathcal{I}})^{-1}\Phi_{\mathcal{I}}^T R$
**until** $c_j \leq \beta$ **or** $\bar{\mathcal{I}} = \{\}$

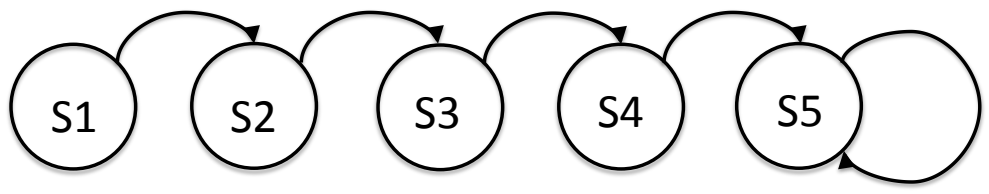

*Figure 1.* A Markov chain for which OMP-TD cannot achieve sparse recovery with an indicator function basis.

which, among features not already in use, currently has the highest correlation with the residual. After adding a feature, the new fixed point is computed using the closed form LSTD fixed point equation, and the new residual is computed. The main results of the BEBF paper apply to OMP-TD: mainly, that each new feature improves a bound on the distance between the fixed point and the true value function $V^*$, as long as the correlation between the feature and the residual is sufficiently large.

### 4.1. Sparse Recovery in OMP-TD

**Theorem 2** *Even if $V^*$ is $m$-sparse in an orthonormal basis, OMP-TD cannot guarantee exact recovery of $V^*$ in $m$ iterations.*

Proof (By counterexample) Consider the Markov chain in figure 1. The arcs indicate deterministic transitions. Suppose $R(S2) = R(S3) = R(S4) = 1$, $R(S5) = 0$, and $R(S1) = -(\gamma + \gamma^2 + \gamma^3)$, then $V^* = [0, 1 + \gamma + \gamma^2, 1 + \gamma, 1, 0]$. With an orthonormal basis $\Phi$ defined by the indicator functions $\varphi_i(s) = I(s = s_i)$, $V^*$ is 3-sparse in $\Phi$, with $opt = \{2, 3, 4\}$. Starting from the empty set of features, the residual vector is just $R$. OMP-TD will pick the vector with the correlation with the residual, which will be $\varphi_1$, a vector that is not in *opt*. □

The next feature added by OMP-TD could be $S4$, then $S3$ and $S2$. Selecting a single vector not in *opt* suffices to establish the proof, which means that we could have shortened the example by removing states $S2$ and $S3$ and connecting $S1$ directly to $S4$. However, the longer chain is useful to illustrate an important point about OMP-TD: It is possible to add a gratuitous basis function at the very first step of the algorithm and the mistake may not be evident until an arbitrary number of additional basis functions are added. This example is easily extended so that an arbitrary number of gratuitous basis functions are added before the first basis function in *opt* is added by making multiple copies of the $S1$ state (together with the corresponding indicator function features). Such constructions can defeat modifications to OMP-TD that use a window of features and discard gratuitous ones (Jain et al., 2011) for any fixed-size window.

An algorithm that chooses features from $\Phi$ based upon the Bellman residual (OMP-BR in the terminology of Johns (2010)), would suffer the same difficulties as OMP-TD in this example. The central problem is that *the Bellman error may not be a trustworthy guide* for selecting features from $\Phi$ even if $\Phi$ is orthogonal.

### 4.2. Sparse Recovery in OMP-BRM

**Lemma 1** *If $V^*$ is $m$-sparse in $\Phi$, then $R$ is at least $m$-sparse in $(\Phi - \gamma P\Phi)$.*

Proof Since $V^* = (I - \gamma P)^{-1}R$, we have

$$\begin{aligned}(I - \gamma P)^{-1}R &= \Phi_{opt} w_{opt}\\ R &= \Phi_{opt} w_{opt} - \gamma P \Phi_{opt} w_{opt}\\ &= [\Phi - \gamma P\Phi]_{opt} w_{opt}.\end{aligned}$$ □

The implication is that we can perform OMP on the basis $(\Phi - \gamma P\Phi)$ and, if there is a sparse representation for $R$ in the basis, we will obtain a sparse representation of $V^*$ as well. This permits a sparse recovery claim for OMP-BRM that is in stark contrast to the negative results for OMP-TD.

**Theorem 3** *If $V^*$ is $m$-sparse in $\Phi$, and*

$$\max_{i\notin opt} \|X_{opt}^{\dagger} x_i\|_1 < 1, \tag{4}$$

*for $X = \Phi - \gamma P\Phi$, then OMP-BRM called with $\Phi$, $\Phi' = P\Phi$, $R$, $\gamma$, and $\beta = 0$ will return $w$ such that $V^* = \Phi w$ in at most $m$ iterations.*

Proof (sketch) The proof mirrors a similar proof from Tropp (2004) and is provided in the detail in the full version of the paper[3]. Lemma 1 implies that $R$ is $m$-sparse in X.

[3]Full version available at `http://www.cs.duke.edu/~parr/icml2012-full.pdf`

Since OMP-BRM does OMP with basis $X$ and target $R$, the sparse recovery results for OMP for regression apply directly. □

Tropp's extension to the approximate recovery case also applies directly to OMP-BRM (see full version of the paper[3]). For the noisy case, we expect that the results of Zhang (2009) could be generalized to RL, but we defer that extension for future work.

### 4.3. Sparse Recovery Behavior

We set up experiments to validate the theory of exact recovery for OMP-BRM, and to investigate the behavior of OMP-TD in similar circumstances. First, we generated a basis for the 50-state chain problem (see section 5) in which the first three basis functions provide an exact reconstruction of $V^*$, and the remaining 997 features are randomly generated, which satisfies equation (4). (Generating such a basis required first generating a much larger (50 x 3000) matrix, then throwing out features which violated the exact recovery condition, and finally trimming the matrix back down to 1000 features. The first three features were constructed by finding two random features which highly correlate with $V^*$, then adding in a third feature which was the reconstruction residual using the first two features.)

By using the resulting basis in OMP-BRM with exact data (i.e., where $\Phi' = P\Phi$), we found that, for sufficiently small threshold value, OMP-BRM uses exactly the first three features in its approximation, in accordance with theory. We also tried OMP-BRM with noisy data by sampling 200 state transitions from the 50-state chain problem. In this case, we found that OMP-BRM reliably selected the first three features before selecting any other features. Interestingly, the same basis proved to enable exact recovery for OMP-TD as well. Using the same 200 samples, OMP-TD selected the first three features before selecting any other features.

## 5. Experiments

While we have theoretical results for OMP-BRM that suggest we should be able to perform optimal recovery under certain conditions on the feature dictionary, practical problems contain noise, which current theory does not address. In addition, the desired conditions on the feature dictionary may not hold and it can be difficult to verify if they do hold since these conditions are a property of both the features and the transition function for OMP-BRM. For OMP-TD, we have negative worst-case results, but Section 4.3 gives hope that things may not be so dismal in practice. To gain some understanding of how these algorithms perform under typical conditions, we performed experiments on a number of benchmark RL problems. Figure 2 shows our main results, the approximation accuracy achieved by each of four algorithms on each problem. Table 1 summarizes the benchmark problem properties and some experimental settings. The algorithms studied included:

*OMP-BRM*: The OMP-BRM algorithm as described above, but with some additional machinery to improve performance in actual use. It is well known (Sutton & Barto, 1998) that BRM is biased in the presence of noise, i.e., when samples are taken from a stochastic transition function. One solution to this problem is to use double samples for each transition. In each of our experiments, we ran with and without doubled samples, and we report the behavior of the better performing option. Table 1 records which experiments use doubled samples. Figure 3 shows how doubling samples affects performance on the 50-state chain problem. In all cases, the number of samples in the *Samples* column refers to the number of starting states.[4] In the doubled case we also add in a small amount (0.01) of $L_2$ regularization in order to keep the algorithm well behaved (by keeping the matrices to be inverted well conditioned).

*OMP-TD*: The OMP-TD algorithm as described above, but with a small amount (0.01) of $L_2$ regularization when computing the final solution at each threshold. This change seemed to be important for some of the harder benchmarks such as puddleworld and two-room, where without regularization, the LSTD solution at various threshold values would occasionally exhibit unstable behavior. We use regularization on all problems, as it seems to cause no issues even for benchmarks which do not need it.

*LARS-BRM*: This is our implementation of the algorithm of Loth et al. (2007), which effectively treats BRM as a regression problem to be solved using LARS. Note that there is no provision for sample doubling in this algorithm, which may affect its performance on some problems.

*LARS-TD*: This is the LARS-TD algorithm of Kolter & Ng (2009). Again we found it useful sometimes to include a small amount (0.01) of $L_2$ regularization, giving the "elastic net" (Zou & Hastie, 2005) solution. The additional regularization does not always benefit; however, we report in figure 2 the better of the two for each benchmark. Table 1 records which experiments use $L_2$ regularization in LARS-TD.

In figure 2, we have plotted performance of the various algorithms on a variety of benchmark problems. The vertical axis is the root mean square error with respect to the true value function, $V^*$. For the discrete state problems, $V^*$ is computed exactly, while for the continuous state prob-

[4]This could be interpreted as giving the double-sample case an unfair advantage because there is no "penalty" for the additional samples collected. On the other hand, counting next states only could be seen as imposing too harsh of a penalty on OMP-BRM since the same number of samples would necessarily have sparser coverage of the state space.

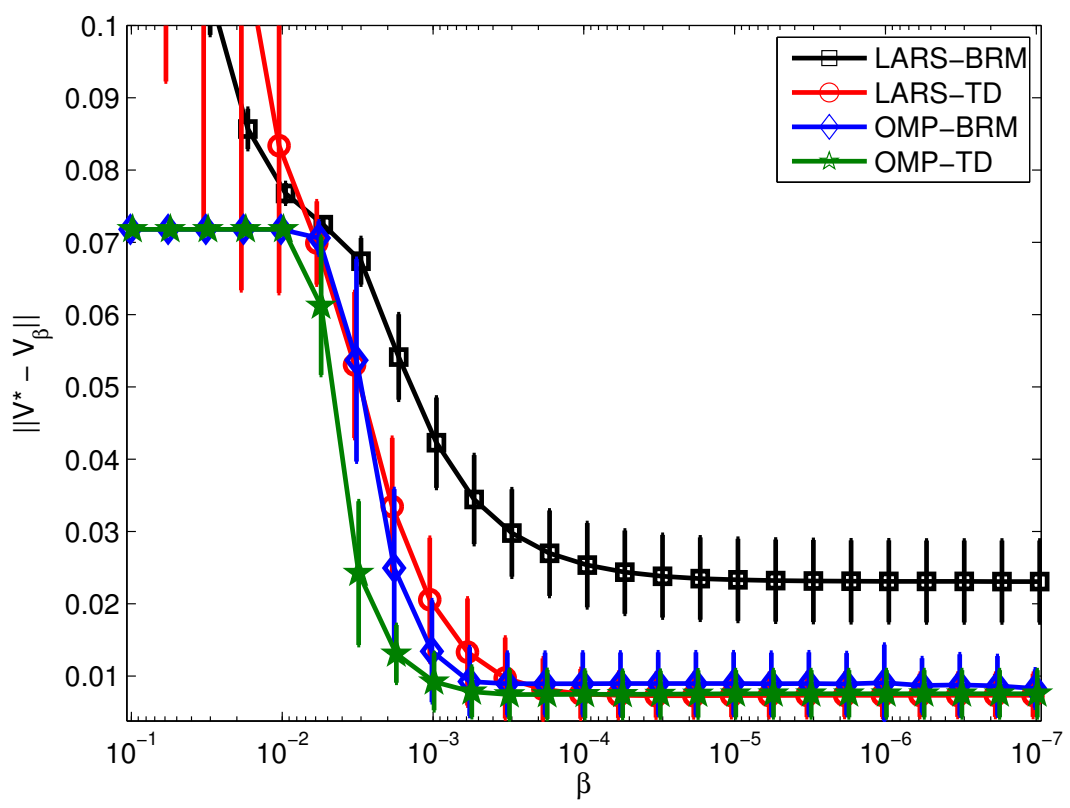

(a) Chain

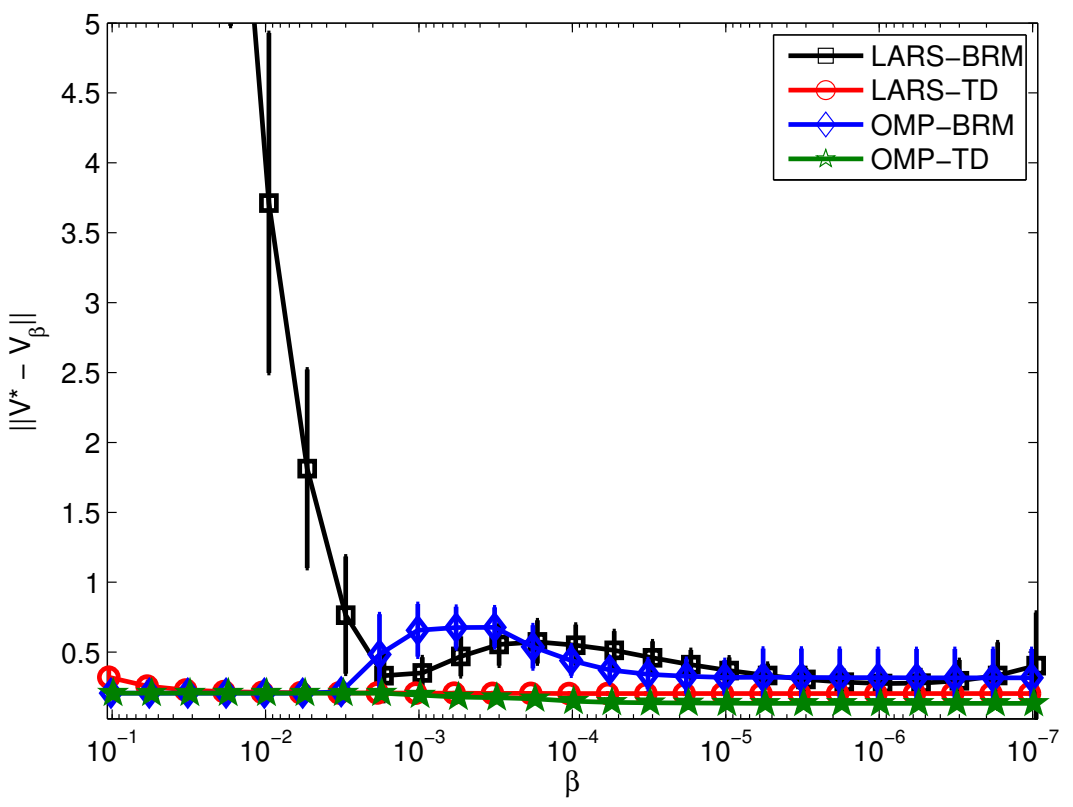

(b) Pendulum

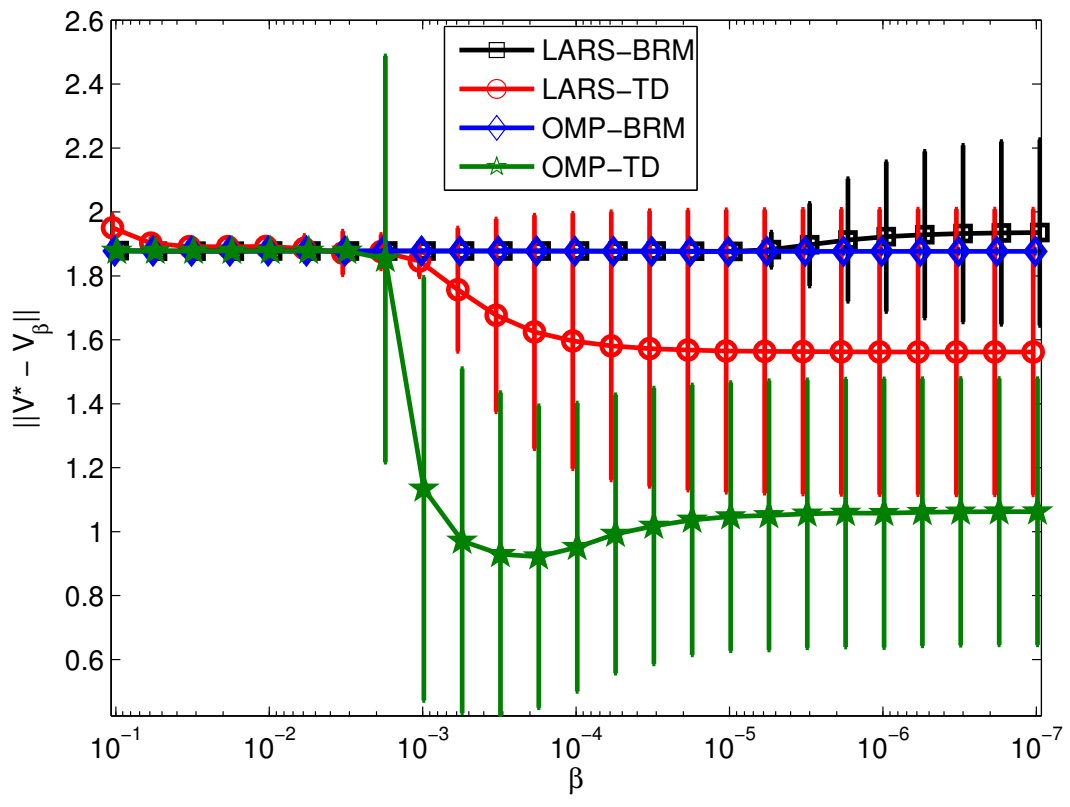

(c) Blackjack

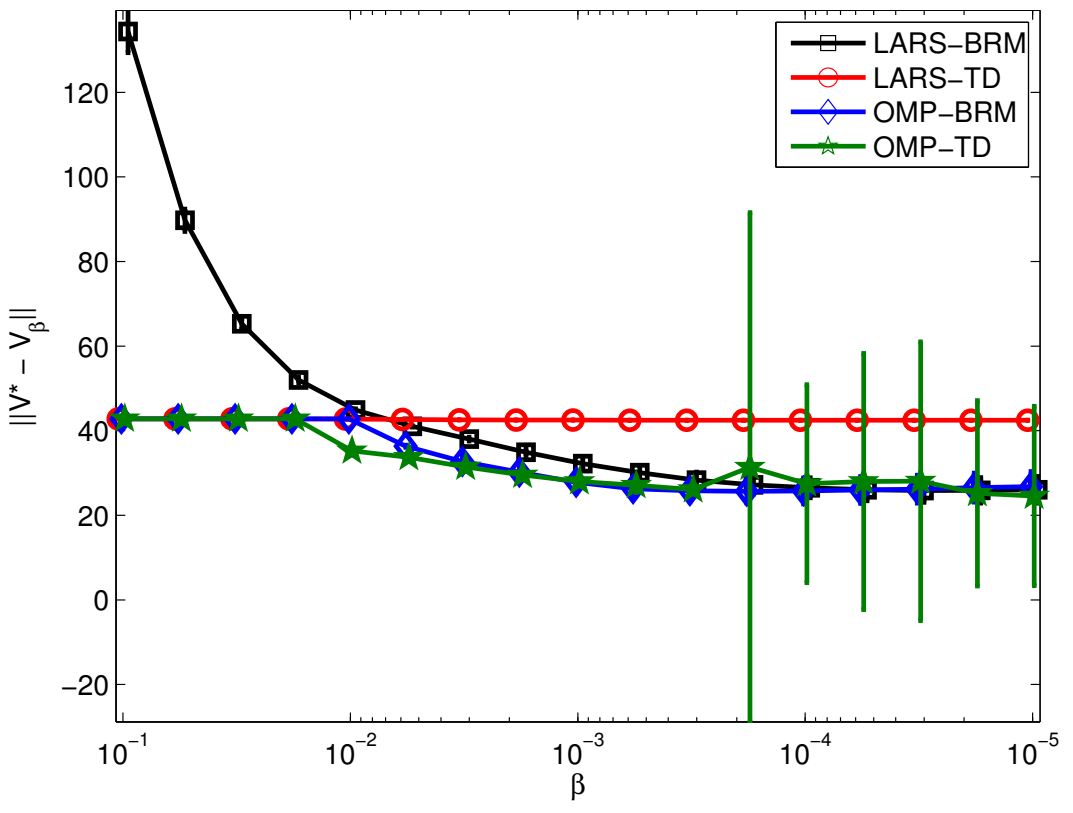

(d) Mountain Car

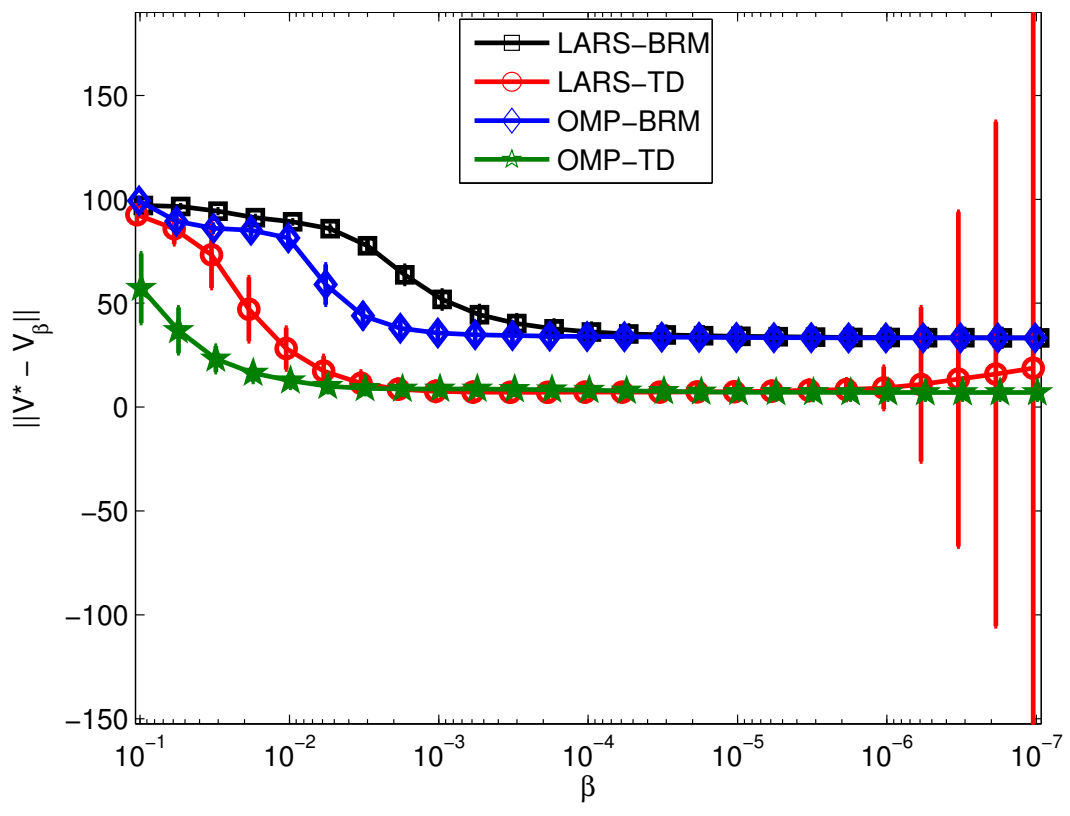

(e) Puddleworld

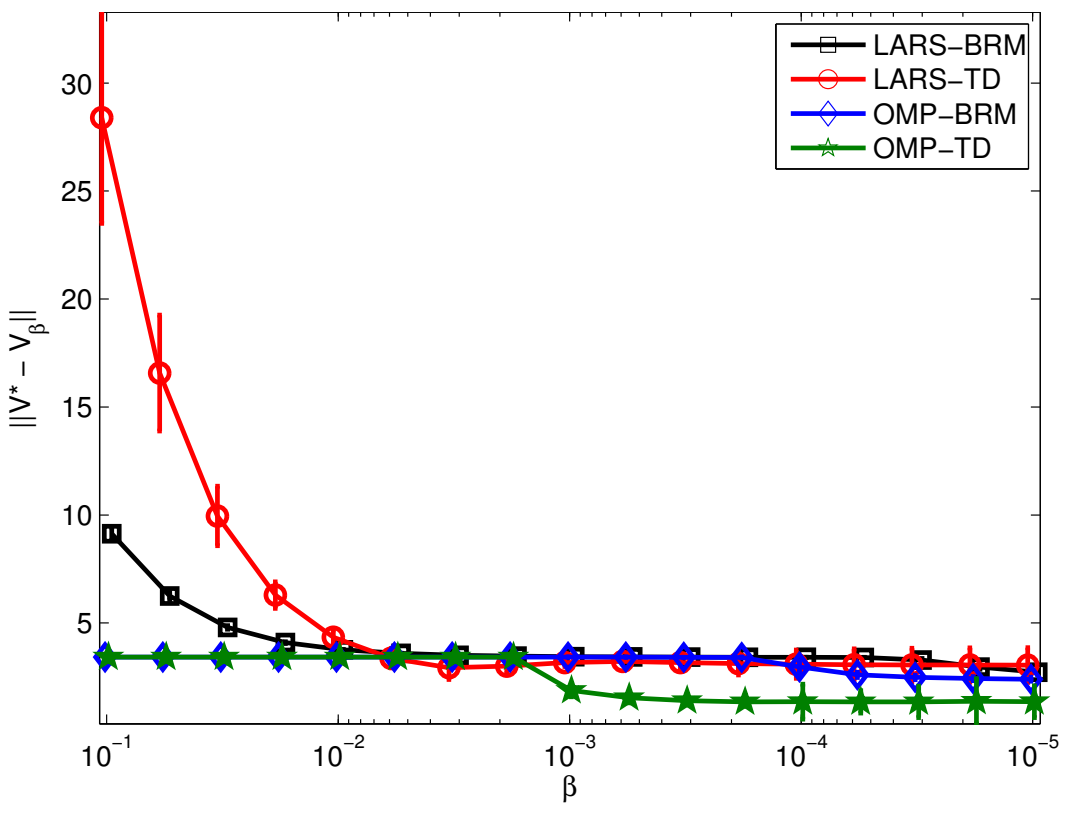

(f) Two Room

*Figure 2.* Error with respect to $V^*$ versus regularization/threshold coefficient on several benchmark problems.

lems $V^*$ is computed at a large number of sampled states by Monte Carlo rollouts. The number of trials over which the values are averaged is reported in table 1. The horizontal axis gives the threshold/regularization coefficient value $\beta$ for the value function $V_\beta$ plotted.

In general, the meaning of $\beta$ for the OMP algorithms is different than for the LARS algorithms. However, there is a strong similarity between the two in that, for both, a solution at value $\beta$ implies that there are no further features with a correlation with the current residual of $\beta$ or larger. This is explicit in the OMP algorithms, and implicit in the fixed point conditions for LARS-TD (n.b. Kolter & Ng (2009), equation 9) and LARS-BRM. Surprisingly, given that OMP is greedy and never removes features, we find that the sparsity of the solutions given by the algorithms is similar for the same values of $\beta$; e.g., see figure 4.

*Table 1.* Benchmark experiment properties and experimental settings.

| Problem | State space | Features | Samples | Trials | LARS-TD $L_2$? | BRM double samples? |
|---|---|---|---|---|---|---|
| Chain | Discrete, 50 states | 208 | 500 | 1000 | × | √ |
| Pendulum | Continuous, 2d | 268 | 200 | 1000 | √ | √ |
| Blackjack | Discrete, 203 states | 219 | 1600 | 1000 | × | × |
| Mountain Car | Continuous, 2d | 1366 | 5000 | 100 | √ | × |
| Puddleworld | Continuous, 2d | 570 | 2000 | 500 | × | × |
| Two Room | Continuous, 2d | 2227 | 5000 | 1000 | × | × |

As figure 2 shows, OMP-TD is generally competitive with, or better than, LARS-BRM and LARS-TD on most of the benchmark problems. In addition, we note that the OMP-based algorithms are considerably faster than the LARS-based algorithms; see figure 5 for a comparison of computation time on the puddleworld problem.

While OMP-TD generally leads in our benchmarks, we should point out some caveats. With very small numbers of samples (even fewer than shown in our experiments) OMP-TD was somewhat more prone to unstable behavior than the other algorithms. This could simply mean that OMP-TD requires more $L_2$ regularization, but we did not explore that in our experiments. An indication that OMP-TD can require more $L_2$ regularization than the other algorithms is evident in our OMP-TD experiments for Mountain Car, where the high variance for low values of $\beta$ arises from just two (out of 100) batches of samples. In these cases, it appears that OMP-TD is adding a feature which is essentially a delta function on a single sample. Without additional $L_2$ regularization, OMP-TD produces very poor value functions for these batches.

For all of the experiments except for blackjack, the features are radial basis functions (RBFs). There are multiple widths of RBFs placed in the state space in grids of various spacing. For blackjack, the basis functions are indicators on groups of states. All features are normalized, although we tried both with and without normalization, and the results were qualitatively similar.

More information about each of the experimental domains can be found in the full version of the paper[3].

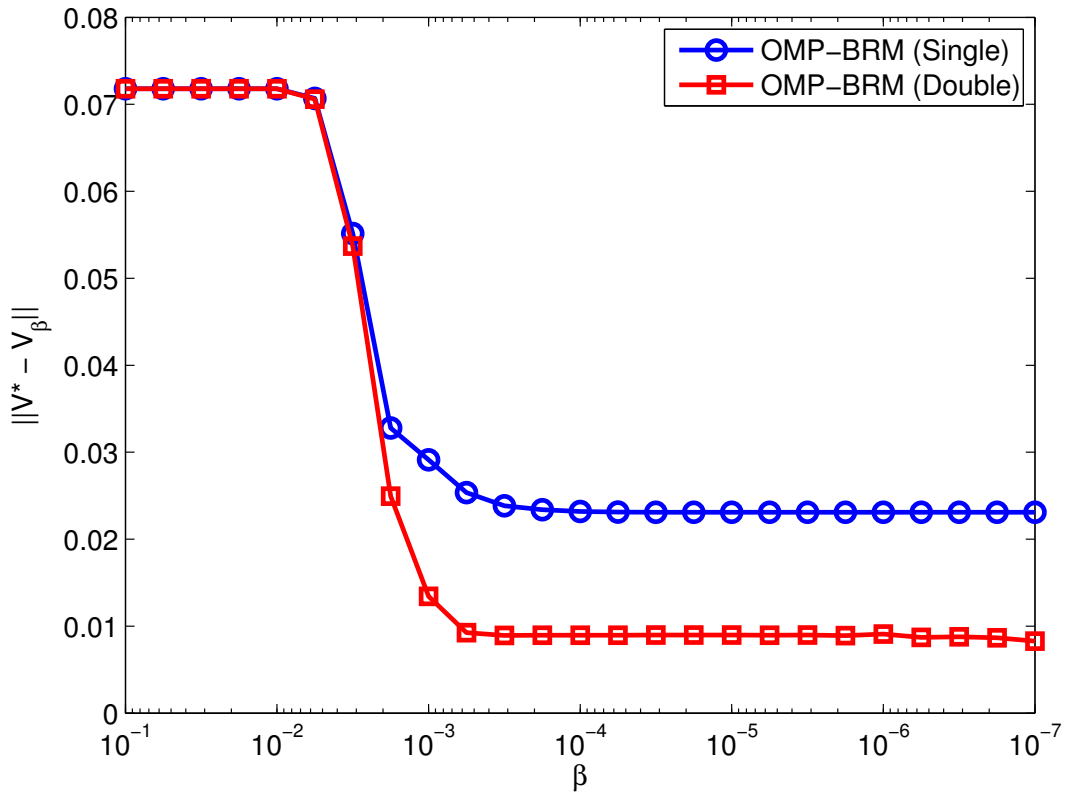

*Figure 3.* Comparison of error between OMP-BRM using single samples versus doubled samples on chain.

## 6. Conclusions and Future Work

In this paper we have explored the theoretical and practical applications of OMP to RL. We analyzed variants of OMP and compared them experimentally with existing $L_1$ regularized approaches. We showed that perhaps the most natural scenario in which one might hope to achieve sparse recovery fails; however, one variant, OMP-BRM, provides promising theoretical guarantees under certain assumptions on the feature dictionary. Another variant, OMP-TD, empirically outperforms prior methods both in approximation accuracy and efficiency on several benchmark problems.

There are two natural directions for further development of this work. Our theoretical results for OMP-BRM built upon the simplest results for sparse recovery in regression and do not apply directly to more realistic scenarios that involve noise. Stronger results may be possible, building upon the work of Zhang (2009). A more interesting, but also more challenging, future direction would be the theoretical development to explain the extremely strong performance of OMP-TD in practice despite negative theoretical results on sparse recovery. For example, there could be an alternate set of conditions on the features that frequently hold in practice and that can be shown theoretically to have good sparse recovery guarantees.

## Acknowledgments

This work was supported by NSF IIS-1147641. Opinions, findings, conclusions or recommendations herein are those of the authors and not necessarily those of NSF. The authors also wish to thank Susan Murphy and Eric Laber for helpful discussions.

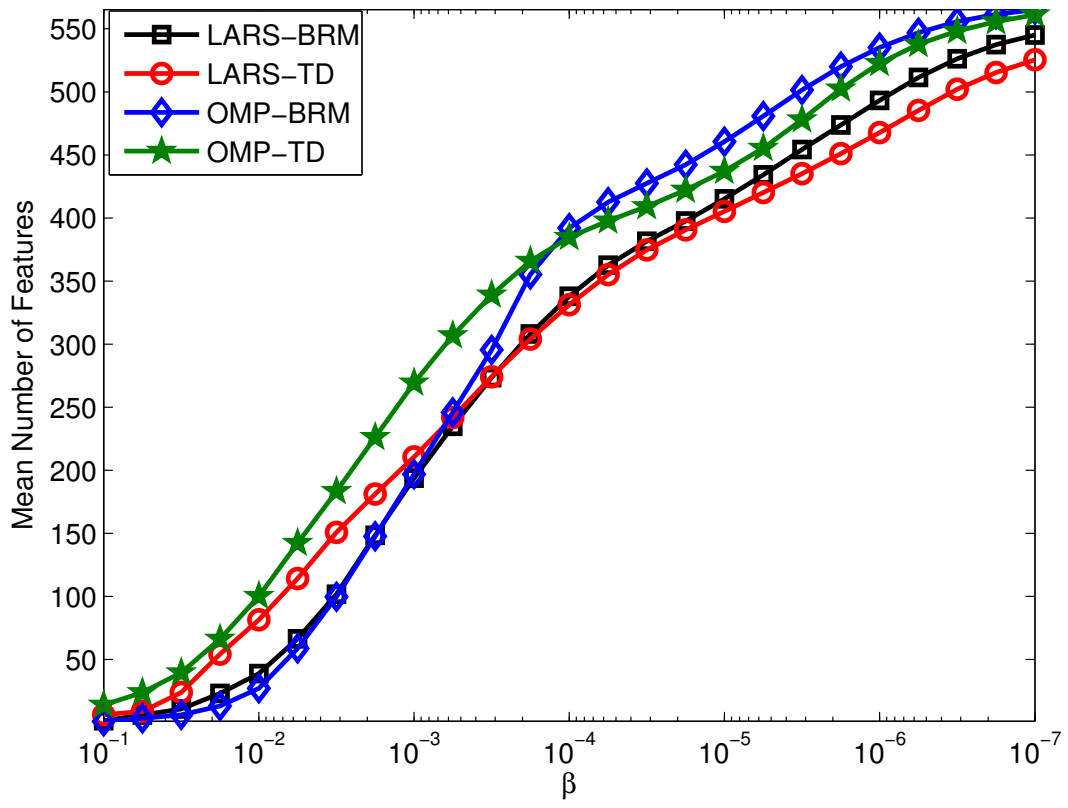

*Figure 4.* Sparsity (mean number of features) versus regularization/threshold coefficient on puddleworld.

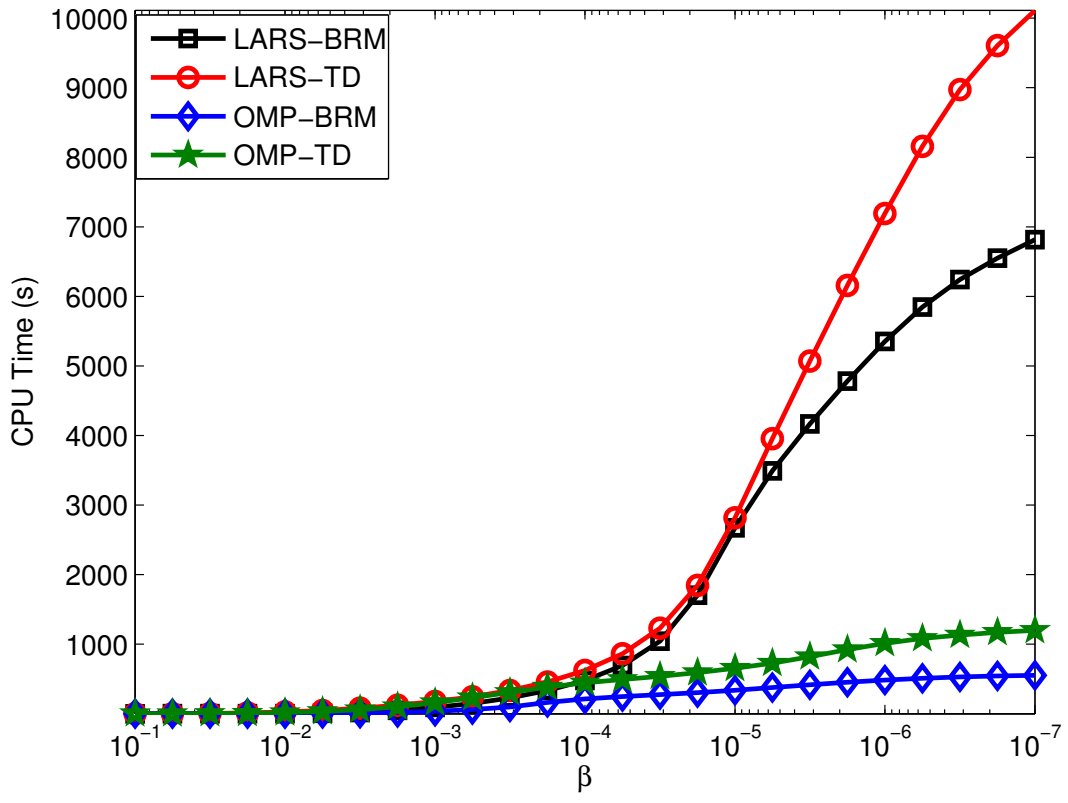

*Figure 5.* Execution time versus regularization/threshold coefficient on puddleworld.